\documentclass[12pt]{article}

\usepackage{siunitx}
\def\bP{\mathbb{P}}
\usepackage{bbm}
\usepackage{dsfont}
\usepackage{array}
\usepackage{xcolor}
\usepackage[normalem]{ulem}
\usepackage{subfigure} 
\usepackage{amsmath} 
\usepackage{amsfonts}
\usepackage{booktabs}
\usepackage{tabularx}

\usepackage{amsthm}
\usepackage{amsmath} 
\usepackage{amsfonts}
\usepackage{amssymb}
\def\bP{\mathbb{P}}
\usepackage{bbm}
\usepackage{dsfont}
\usepackage{graphicx}
\usepackage{hyperref}

\usepackage{times}

\newcounter{lastnote}

\begin{document} 
\title{On a scalable entropic breaching of the overfitting barrier in machine learning
}
\author
{ Illia Horenko$^{1,\ast}$\\
\small{$^{1}$Universit\`a della Svizzera Italiana, Faculty of Informatics, Via G. Buffi 13, TI-6900 Lugano, Switzerland,}\\
\small{$^\ast$To whom correspondence should be addressed; E-mail: horenkoi@usi.ch }\\}

\maketitle
\begin{abstract}
Overfitting and  treatment of  "small data" are among the most challenging problems  in the machine learning (ML), when a relatively small data statistics size $T$ is not enough to provide a robust ML fit for a relatively large data feature dimension $D$.  Deploying  a massively-parallel ML analysis of generic classification problems for different $D$ and $T$, existence of statistically-significant  linear overfitting barriers for common ML methods is demonstrated. For example, these results reveal that for a robust classification of bioinformatics-motivated generic problems with the Long Short-Term Memory deep learning classifier (LSTM) one needs  in a best case a statistics $T$ that is at least 13.8 times larger then the feature dimension $D$. It is shown that this overfitting barrier can be breached at a $10^{-12}$ fraction of the computational cost by means of the  entropy-optimal Scalable Probabilistic Approximations algorithm (eSPA), performing a joint solution of the entropy-optimal Bayesian network inference and feature space segmentation problems. 
  Application of eSPA to experimental single cell RNA sequencing data  exhibits a 30-fold classification performance boost when compared to standard bioinformatics tools - and a 7-fold boost when compared to the deep learning LSTM classifier.    
\end{abstract}

\baselineskip24pt 

\section*{{\emph{Introduction}}}
Despite of the numerous success stories of data science and deep learning methods reported in the recent years,  their robustness remains an issue of a major concern. A broad range of data science problems - for example in biomedical research and in economics - are characterised by the number of data features $D$ that can be many orders of magnitude larger then the available data statistics $T$. Applying advanced machine learning tools with multiple tuneable parameters $N$  to such "small data problems"  typically results in what is called an overfitting problem - when a very good model fit on the training data does not lead to a good prediction on the validation data \cite{hawkins04,clauset19}. The "small data challenge" and the overfitting problems attract an increasing attention across data science domains \cite{clauset18,clauset19,han19,dsouza20} - and very recent theoretical findings indicate that the deep learning methods are less sensitive to overfitting, due to the implicit numerical regularisation and "zeroing-out" of the irrelevant tuning parameters \cite{poggio17}.

However, a systematic evaluation and comparison of various ML methods in a space spanned by the statistics size $T$, feature dimension $D$ and model complexity $N$ is hampered by their high computational cost - and by the quick growth of this cost with $T$, $D$ and $N$ \cite{bottou10,Bengio2012,deep_comp_bio16} . 

This manuscript features two very simple toy model systems motivated by a realistic "small data" problem from the bioinformatics - where a distinction between the two classes of subjects (long-living C. Elegans worms and worms with a "normal" life span) is only achieved when taking into account an orchestrated  change in two particular longevity-related gene expression levels - and where the remaining genes do not play any role in this distinction \cite{lan19} . 

\begin{figure*} 
\begin{center} 
     \includegraphics[width=1.0\textwidth]{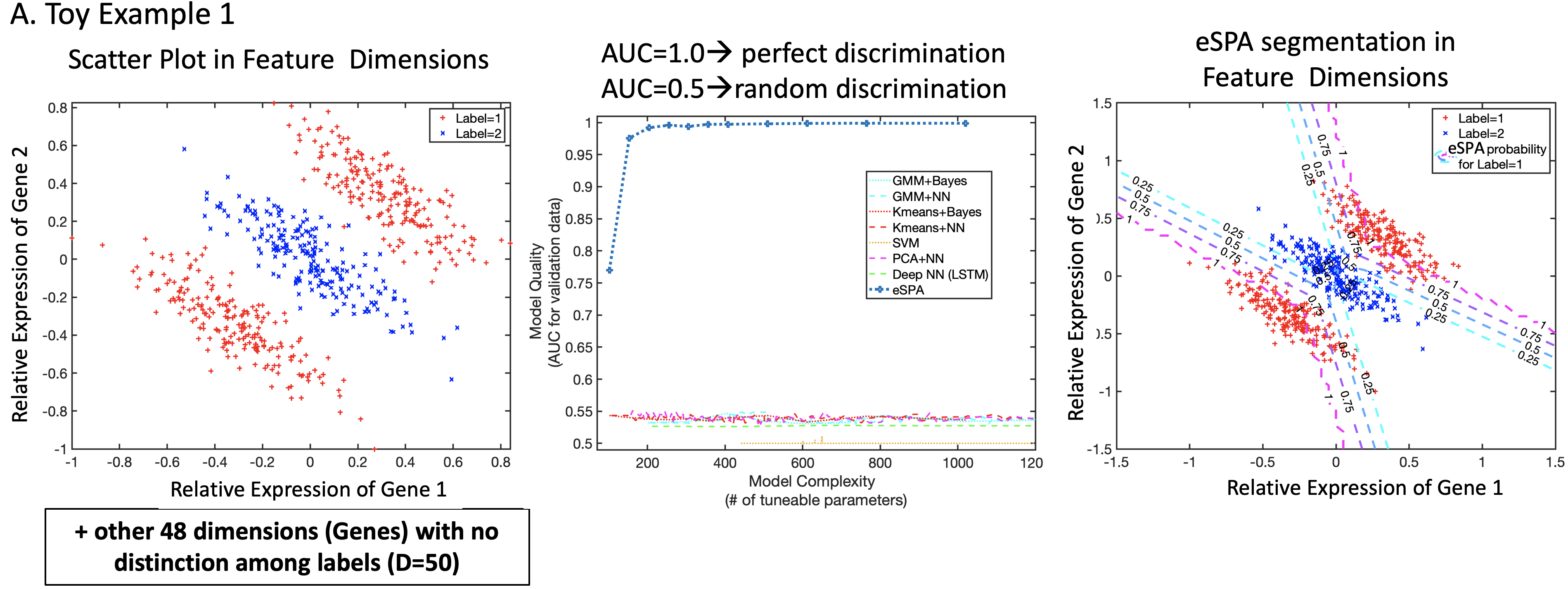} 
     \includegraphics[width=1.0\textwidth]{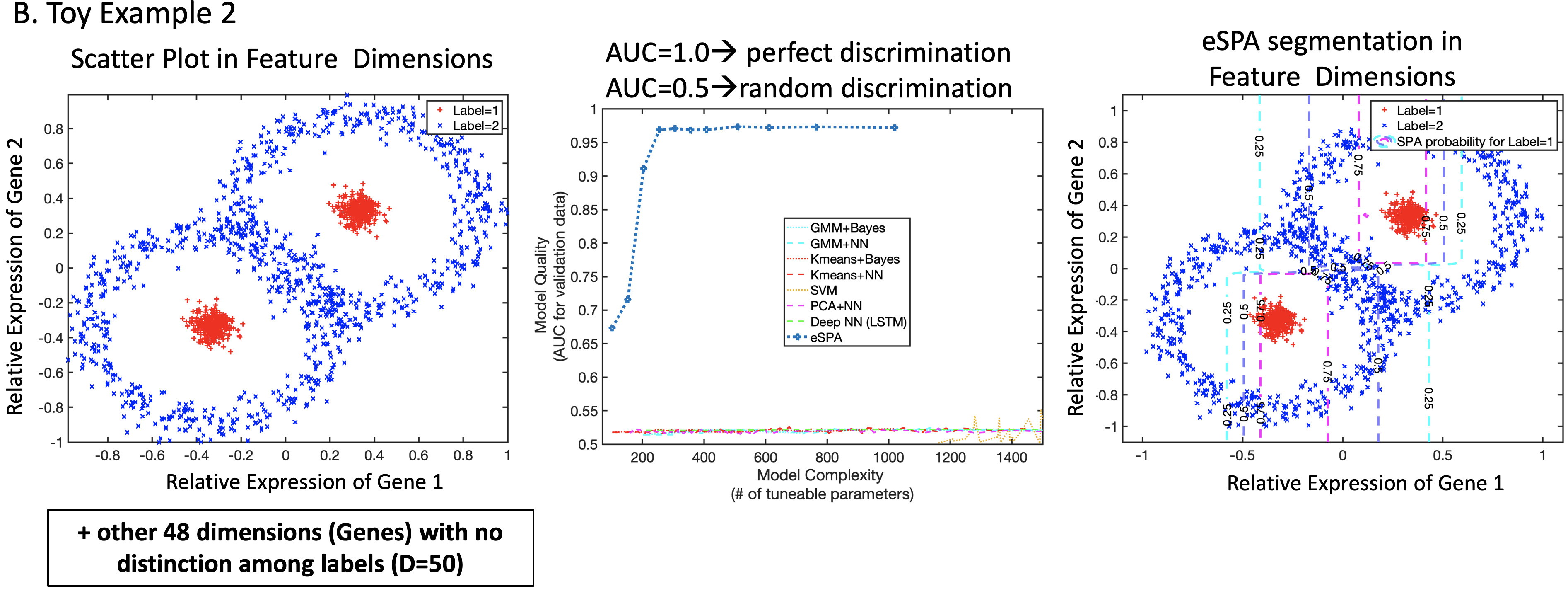} 
      \includegraphics[width=1.0\textwidth]{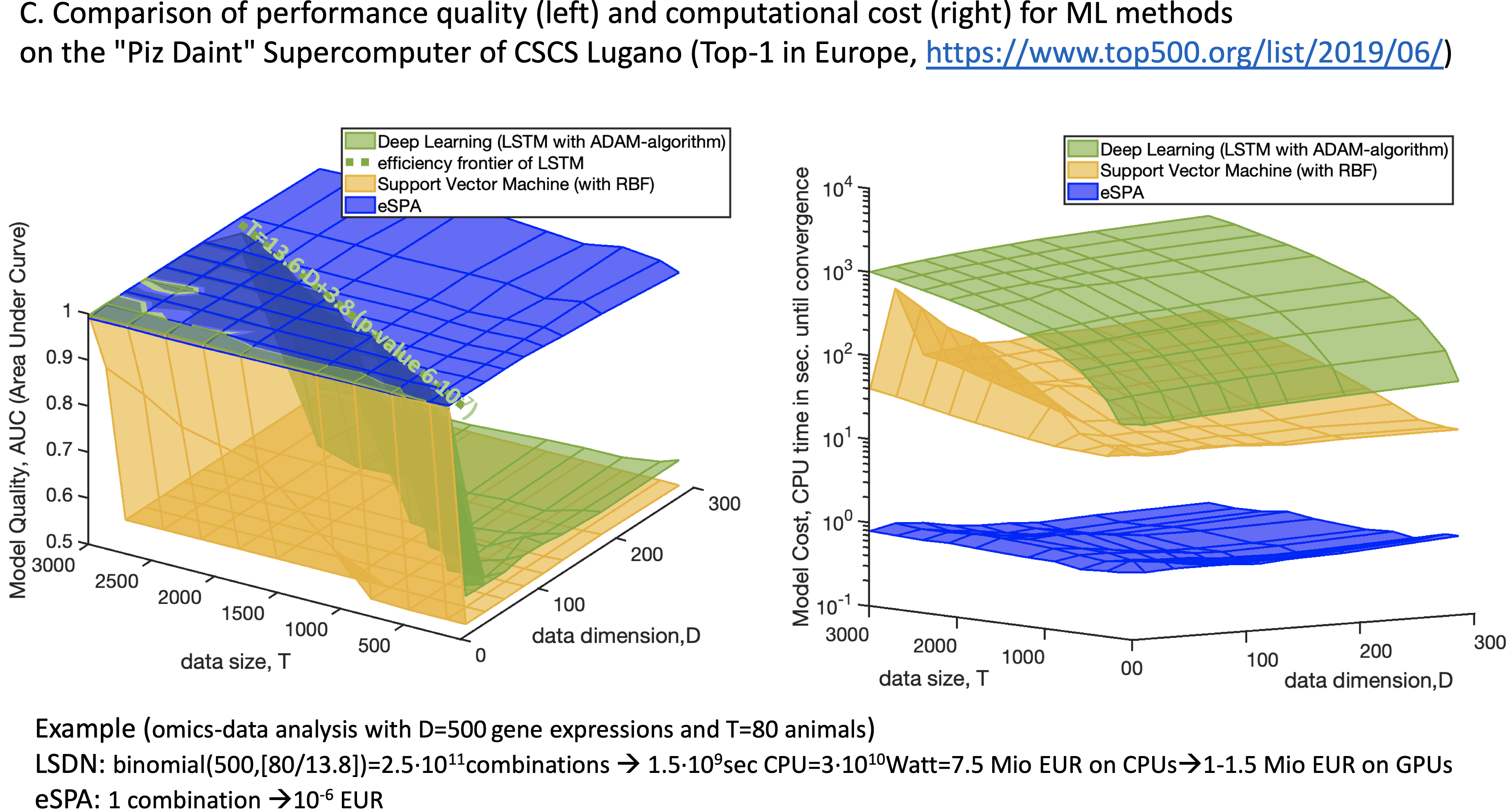} 
\end{center}
   \caption{
  {\bf  Comparison of common Machine Learning (ML) tools to the entropy-based Scalable Probabilisitc Approximation Algorithm (eSPA)  for  bioinformatics-motivated model examples.}
   }
    \label{fig:toy}
\end{figure*}

Using random data outputs from these generic toy systems, multiply cross-validated comparisons of classification performances and computational costs for common machine learning methods (including Bayesian methods, support vector machines, as well as shallow, reinforced and deep neuronal networks) is then  performed on the "Piz Daint" supercomputing facility of the Swiss Supercomputing Center in Lugano (top-1 in Europe).
 Results of this comparison reveal that the common machine learning methods  exhibit a statistically-significant linear overfitting barrier in the space of $D$ and $T$, indicating that, for example, Long Short-Term Memory deep learning networks (LSTM) would in the best case require at least 13.8 times more subjects $T$ then the involved gene feature dimensions $D$ in order to reveal the correct labelling rule behind these toy models (see Fig.~\ref{fig:toy}). 
 
 Next, an entropy-based generalisation of the recently introduced Scalable Probabilistic Approximation (SPA) method \cite{gerber20} is presented. Resulting entropy-based SPA (eSPA) seeks for as simple as possible (but not simpler then necessary) geometric segmentations of the given labelled data into $K$ geometric boxes in the feature dimensions, simultaneously  optimising a fit of the Bayesian relationship connecting the probabilities to be in segmentation boxes and the probabilities to attain certain labels (see Fig.~\ref{fig:algorithm_and_sketch1} and Fig.~\ref{fig:algorithm_and_sketch2}).  It is proven that the entropy-optimal Euclidean segmentation in labelled data analysis problems  is given by the piecewise-linear geometric boxes in the feature space. It is also proven that this piecewise-linear geometric box segmentation of the feature space can be approximated numerically by means of the eSPA algorithm with a linear scaling of iteration complexity.  Comparison to common tools on the same generic labelled data analysis examples shows that the eSPA allows to overcome the overfitting barrier, providing robust cross-validated label predictions also in the "small data" regimes (when $T<D$), at a fraction of the computational cost (see Fig.~\ref{fig:toy}). Finally, methods comparison for the single cell RNA-sequencing data from \cite{pollen14}  reveals an around 30-fold classification performance boost of eSPA with respect to the common shallow and reinforced learning tools - and a 7-fold classification improvement when compared to the architecture-optimized deep learning LSTM classifier.

\section*{{\emph{Results}}}
\subsection*{{\emph{ML performance comparison with generic toy model examples from bioinformatics}}}
Bioinformatics and omics data analysis applications are particularly challenging for ML methods not only because of the large gene (feature) dimensions $D$ and relatively-small data statistics T, but also due to the difficulties they impose on common dimension reduction and feature selection tools. For example,  very recently it was shown that the life span of the \emph{Caenorhabditis elegans} worm can be expanded by 500 percent due to the synchronised and correlated expression of a particular pair of genes - and with no particular change of expression for all other genes \cite{lan19}. Common  ML feature selection methods like the linear  Principal Components Analysis (PCA) \cite{joliffe02}   - as well as its nonlinear extensions like the t-stochastic nearest neighbour embedding (t-SNE) \cite{maaten08} - fail to detect the relevant pair of genes in these data.   

One of the simplest synthetic toy model systems capturing these features of the C.elegans longevity data set from  \cite{lan19} is exemplified in the Fig.~\ref{fig:toy}. The group of long-living worms is represented by the Gaussian distribution of blue crosses (each cross is a  "long-living" animal). In the scatter plot spanning the two relevant gene dimensions this group is "sandwiched" between the two Gaussian groups of worms with a "normal" life-span (marked with red pluses, every red plus is a worm with a "normal" life span).  The remaining $(D-2)$ dimensions  are modelled as the uniformly-distributed data samples of the same variance and with no distinction between the groups. 

Applications of PCA and t-SNE to the output of this toy model system with $T>200$ fail to detect the correct pair of genes/features if  $D\geq20$. As demonstrated in the middle panel of the Fig.~1A, for $D=50$ and $T=600$,  common shallow, reinforced and deep ML methods fail to detect the correct rule distinguishing the two groups. Their Area Under Curve (AUC) performance measure takes values around 0.5 on the validation data, practically indicating a completely random classification. This figure also shows that the performance of ML methods on these data does not improve with increasing model complexity and with the growing number of tuneable parameters, changing network architectures and other settings. Also the common statistical tools and the Genome-Wide Association Studies (performing $D$ one-dimensional tests to detect a statistically-significant difference of the marginal mean values between the two groups) fail to detect the relevant dimensions because the marginal distributions of the two groups in the two relevant feature dimensions have the same mean. 

Comparison of  performances and total computation costs  of ML methods on randomly-generated outputs of this toy model system is shown in the Fig.~1C. For every particular combination of $D$ and $T$, 100 random data outputs of the toy model from the Fig.~1A are generated. Each of the generated data sets is randomly subdivided into the  75\% of data for model fitting and 25\% of validation data that is not used in the model training. Each of the ML model types is then trained and validated for various settings of possible intrinsic configurations (various network architectures, numbers of hidden neurons, batch sizes), according to the general hyper-parameter selection guidelines provided in the literature \cite{Bengio2012,deep_comp_bio16}. Every point in the surfaces from Fig.~1C represents a mean value over  the 100 optimized models for each of the considered ML methods and for each of the combinations of the data size $T$ and the feature dimension $D$. 

As can be seen from the left panel of Fig.~1C, both the SVM and the deep learning LSTM classifiers exhibit sharp AUC barriers in the space spanned by $D$ and $T$. These barriers are linear with a high statistical-significance (p-value=$6\cdot10^{-7}$ for LSTM) and separate the problem domains with AUC values close to 1.0 (where the classifiers detect the correct rule distinguishing between the blue and the red data instances) and the AUC close to 0.5 (where no correct rule is identified).  The deep learning LSTM classifier is more robust than the SVM and allows for the same data  a correct pattern detection for  problems with much larger dimensionality. However, as can be seen from the right panel of the Fig.~1C, this better classification performance of LSTM comes with an around ten-fold increase of the total computational cost - and with a faster polynomial growth of this cost with an increasing $D$ and $T$. These results reveal that even with the most optimal choice of the hyper-parameters (batch size, network architecture, number of hidden neurons, etc.), correct detection of the classification pattern requires at least 13.8 times more data points $T$ than the feature dimensions $D$. For typical bioinformatics and omics data analysis applications with $D$ being much larger than $T$ these results imply that there exists a maximal number of  features that can be used for a robust detection of patterns. This estimate for a maximal number of features is expressed as $D_{\text{max}}=[T/13.8]$ (where $[x]$ means an operation of finding the largest integer being smaller than $x$). Then, the only possibility to identify the correct and robust classification rule would be to make a complete combinatorial search in a set of all possible  $D_{\text{max}}$-long feature combinations taken from a set of $D$ features. The total number of classification problems to be solved with the LSTM is then given by the binomial coefficient $binomial(D,D_\text{max})=D!/(D_\text{max}!(D - D_\text{max})!)$. To give an example, for data from \cite{lan19} (with around 80 animals and 500 gene expressions)  $D_{\text{max}}=5$ and   $binom(D,D_\text{max})=2.5\cdot10^{11}$ LSTM model runs. Taking into account the average total computation time for the single LSTM problem of this size and dimension (the green surface in the right panel of the Fig.~1C), scrambling through all variants  for classifier patterns will cost around $1.5\cdot10^{9}$ CPU seconds - or $3\cdot10^{10}$ Watt of electricity. This is the amount of electricity that is produced in around three hours by a large nuclear power plant with four running reactors and would cost around 7.5 Mio EUR. Using Graphical Processing Units (GPUs) instead of CPUs would allow to reduce this cost to around  1-1.5 Mio EUR. 

In other words, existence of the statistically-significant overfitting frontier from the left panel of the Fig.~1C induces major combinatorial, computational and financial bottlenecks when applying existing ML instruments to the problems similar to the toy problem from the Fig.~1A. An alternative ML approach aiming at overcoming this problem is introduced and evaluated below. It dwells on a generalisation of the recently introduced Scalable Probabilistic Approximation framework \cite{gerber20} towards a case of the labelled data segmentation and a Shannon-entropy \cite{Jaynes1957a} optimisation for the feature selection part.           

\begin{figure*} 
\begin{center} 
    \includegraphics[width=1.0\textwidth]{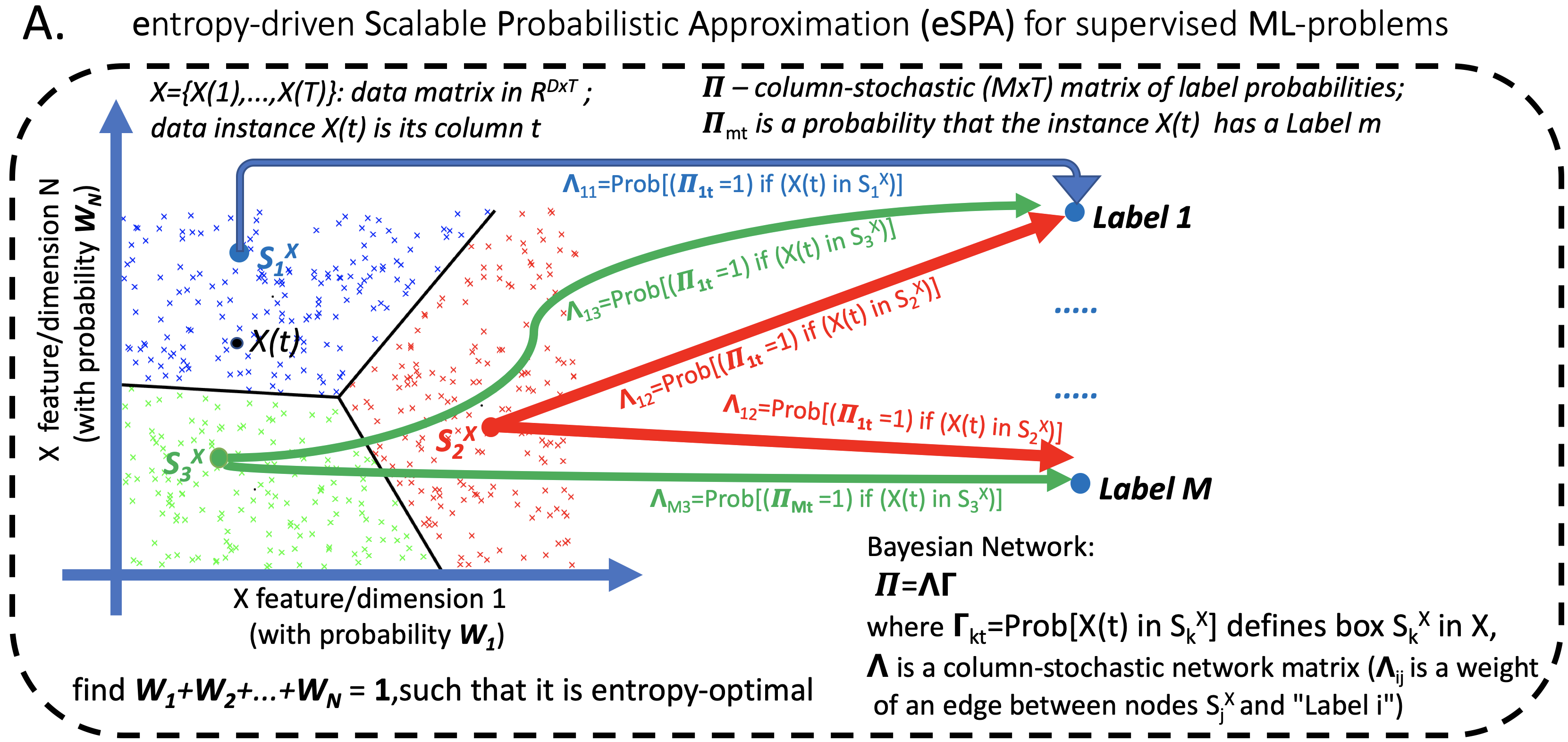}  
     \includegraphics[width=1.0\textwidth]{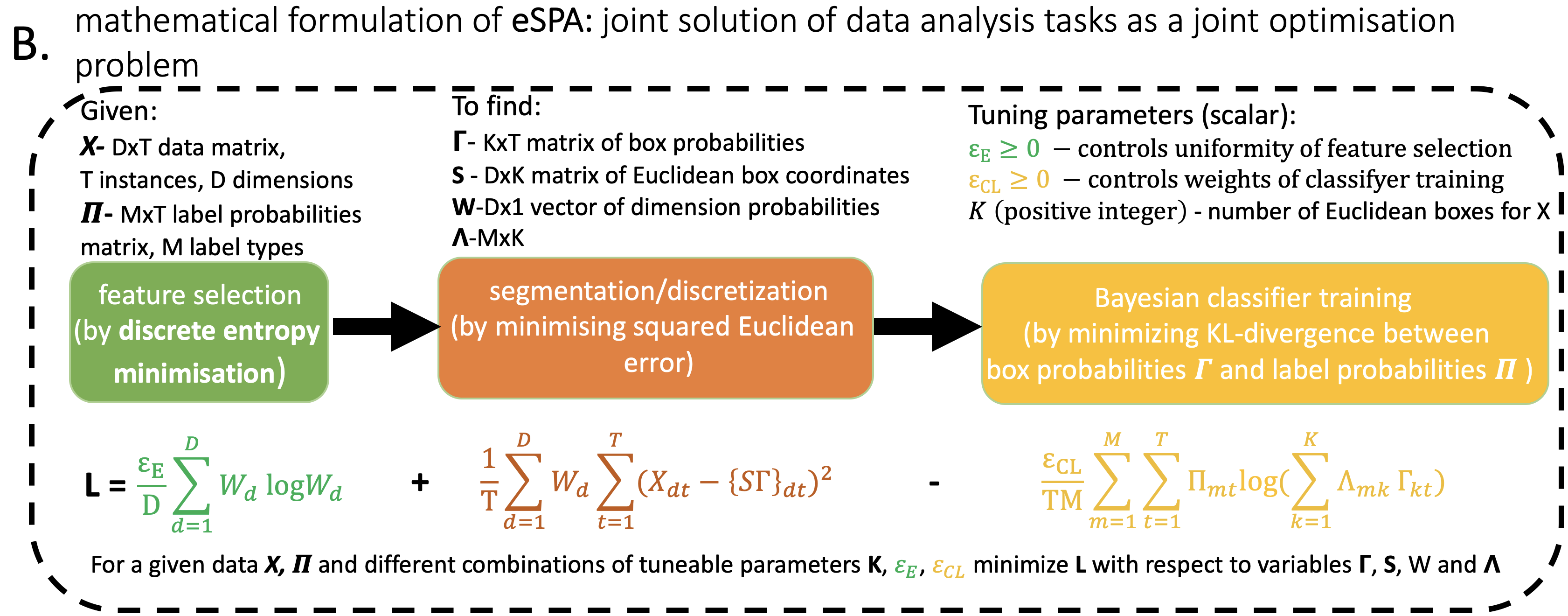} 
 \end{center}
   \caption{ {\bf  Entropy-based Scalble Probabilistic Approximation (eSPA) searches for the simple simultaneous solutions of feature detection, data segmentation and network inference problems}.
   }
    \label{fig:algorithm_and_sketch1}
\end{figure*}  
\subsection*{{\emph{Entropic formulation of the Scalable Probabilistic Approximation problem (eSPA)}}}
Let $X=\left\{X(1),\dots,X(T)\right\}$ be a $D$-times-$T$ real-valued data matrix. Scalable Probabilistic Approximation (SPA)  \cite{gerber20} allows finding a linearly-scalable optimal geometric discretisation/segmentation of the data space by means of $K$ geometric boxes. It is defined in terms of $D$-dimensional box coordinate vectors $S_1^{X},\dots,S_K^{X}$ and a $K$-times-$T$-matrix of box probabilities $\Gamma_{kt}=\bP\left[X(t)\text{ is in }\right.$ $\left.\text{the box }S_k^X\right]$. $\Gamma$ is the stochastic matrix, i.e., $\Gamma\geq0$ and all of its columns sum-up to one. As proven in the Theorems 1 and 3 from \cite{gerber20}, problem of finding an optimal box discretization $\left\{ S,\Gamma\right\}$ (where $S=\left\{S_1^{X},\dots,S_K^{X}\right\}$ is a $D$-times-$K$ matrix of box coordinates) can be solved with a linear iteration complexity scaling simultaneously with an approximate solution of the feature selection problem, by minimizing the regularized squared Euclidean segmentation error:     
\begin{eqnarray}
\label{eq:L_SPA}
L_{\text{SPA}}&=&\frac{1}{TD}\sum_{d=1}^D\sum_{t=1}^T(X_{dt}-\left\{S\Gamma\right\}_{dt})^2 +\epsilon_S\sum_{d=1}^D\sum_{k_1,k_2=1}^K(S_{dk_1}-S_{dk_2})^2.\\
&&\text{such that } \Gamma \text{ is a stochastic matrix.}\nonumber
\end{eqnarray}
As proven in the Theorem 3 from \cite{gerber20}, increasing the tuneable scalar parameter $\epsilon_S\geq0$ minimizes the second term of the above expression. This term provides an upper bound approximation of the box discretization uncertainty - and "zeroes-out" the feature dimensions that are irrelevant for the box discretization. However, the discretization uncertainty approximation provided by the Theorem 3 from \cite{gerber20} and by the second term of the above expression (\ref{eq:L_SPA}) may be too rough and approximate, failing to extract exclusively the feature dimensions that are relevant for finding the most optimal box discretizations $\left\{ S,\Gamma\right\}$. Moreover, SPA is solving the discretization problem in Euclidean space and does not handle the supervised labelled data problems.
  
In the case of supervised ML problems, a $D$-times-$T$ feature matrix $X$ is augmented with the $M$-times-$T$ stochastic matrix of label probabilities $\Pi_{mt}=\bP\left[X(t)\text{ has a label with an index }m\right]$. If the feature data matrix is discretized with boxes $\left\{ S,\Gamma\right\}$,  then the exact Bayesian relation between the probabilities $\Gamma_{kt}$  of a data instance $X(t)$ to be in a certain box $S_k^X$  - and the probabilities $\Pi_{mt}$ for this data instance to have a label with an index $m$ - can be expressed with an exact law of the total probability \cite{Gardiner2004}.  This relationship is provided by an expression $\Pi=\Lambda\Gamma$,  with  a matrix of conditional probabilities $\Lambda_{mk}=\bP\left[X(t)\text{ has a label with an index }m \right.$ $\left.\text{ when/if }X(t)\text{ is in}\right.$ $\left.\text{the box }S_k^X\right]$. These relations between the boxes and the labels can be represented as the Bayesian network (see Fig.~\ref{fig:algorithm_and_sketch1}A). Bayesian $M$-times-$K$ matrix $\Lambda$ is stochastic and, for fixed $\Pi$ and $\Gamma$ it can be found minimizing with respect to $\Lambda$ the average Kullback-Leibler divergence $L_{\text{KL}}$ between $\Pi$ and  $\Lambda\Gamma$: 
\begin{eqnarray}
\label{eq:L_KL}
L_{\text{KL}}&=&-\frac{1}{TM}\sum_{m=1}^M\sum_{t=1}^T\Pi_{mt}\log\left(\sum_{k=1}^K\Lambda_{mk}\Gamma_{kt}\right),\\
&&\text{such that } \Lambda \text{ is a stochastic matrix.}\nonumber
\end{eqnarray}
Such Bayesian and Markovian network inference and community detection methods are widely used in the network science and in various application areas \cite{schuette01,schuette13,clauset15,clauset16,schuette18}. 

Next, the SPA problem (\ref{eq:L_SPA}) for discretization and feature selection is generalized, introducing a vector of $D$ feature probabilities $W$ ($W_i\geq0$ and $\sum_{d=1}^DW_d=1$), where $W_d$ define probabilities of the particular feature dimensions $d$ to contribute to the overall  box-discretization error. Then, constrained minimization of the functional $L_{\text{e}}$ can be written as:
\begin{eqnarray}
\label{eq:L_e}
L_{\text{e}}&=&\frac{1}{T}\sum_{d=1}^DW_d\sum_{t=1}^T(X_{dt}-\left\{S\Gamma\right\}_{dt})^2 +\frac{\epsilon_e}{D}\sum_{d=1}^DW_d\log(W_d),\\
&&\text{such that } \Gamma \text{ is a stochastic matrix and } W \text{ is a probability vector,}\nonumber
\end{eqnarray}
simultaneously minimizing the expected box discretization error (where the expectation is taken with respect to the feature probabilities $W$) in the first term and maximising an entropy of this feature distribution vector $W$ in the second term of the right-hand side in (\ref{eq:L_e}). It is  straightforward to verify that with $\epsilon_e\to\infty$ solutions $W$ converge to the uniform distribution $W_i=1/D$ and that (\ref{eq:L_e}) converges to  the SPA problem  (\ref{eq:L_SPA}) for $\epsilon_S=0$. In other words, SPA method (\ref{eq:L_SPA}) introduced in \cite{gerber20} is a maximum-entropy limit of the more general problem formulation (\ref{eq:L_e}).   For $\epsilon_e=0$ with a fixed box discretization $\left\{ S,\Gamma\right\}$, optimal solution $W$ of (\ref{eq:L_e}) can be computed analytically as $W_d=1/|\kappa|$ if $d\in\kappa$ and $W_d=0$ if $d\notin\kappa$, where $\kappa=\arg\min_{d'}(\sum_{t=1}^T(X_{d't}-\left\{S\Gamma\right\}_{d't})^2)$ and $|\kappa|$ is the number of elements in an index set $\kappa$. In other words, $\epsilon_e\to0$ minimizes an entropy of the feature selection problem and shrinks the set of features to the subset that provides a better discretization with the minimized discretization errors. Scrambling the tuneable parameter $\epsilon_e$ allows to find the entropy-optimal distribution of feature dimensions and zeroes-out the features that are not relevant for the box discretization.

\begin{figure*} 
\begin{center} 
     \includegraphics[width=1.0\textwidth]{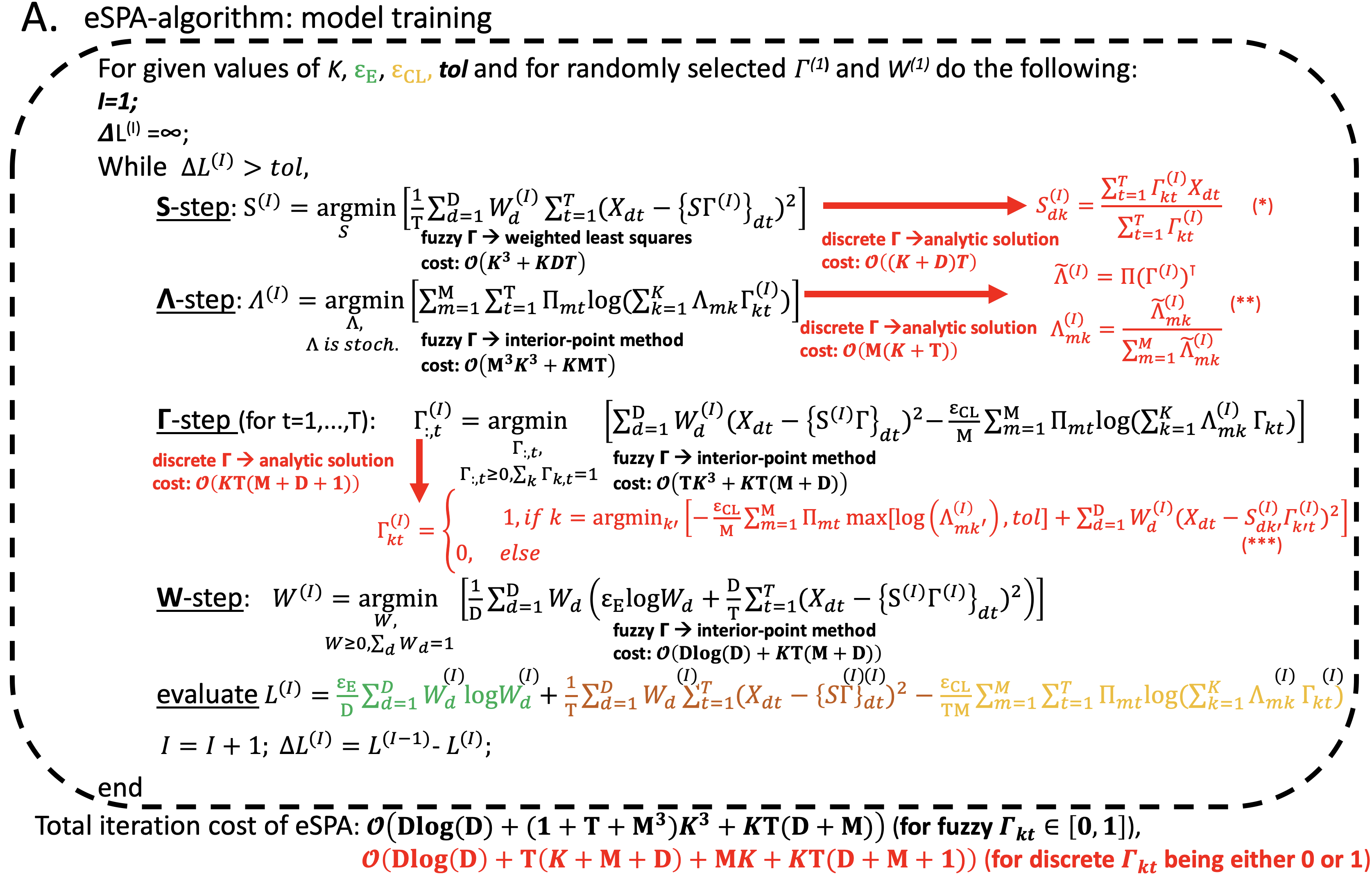} 
     \includegraphics[width=1.0\textwidth]{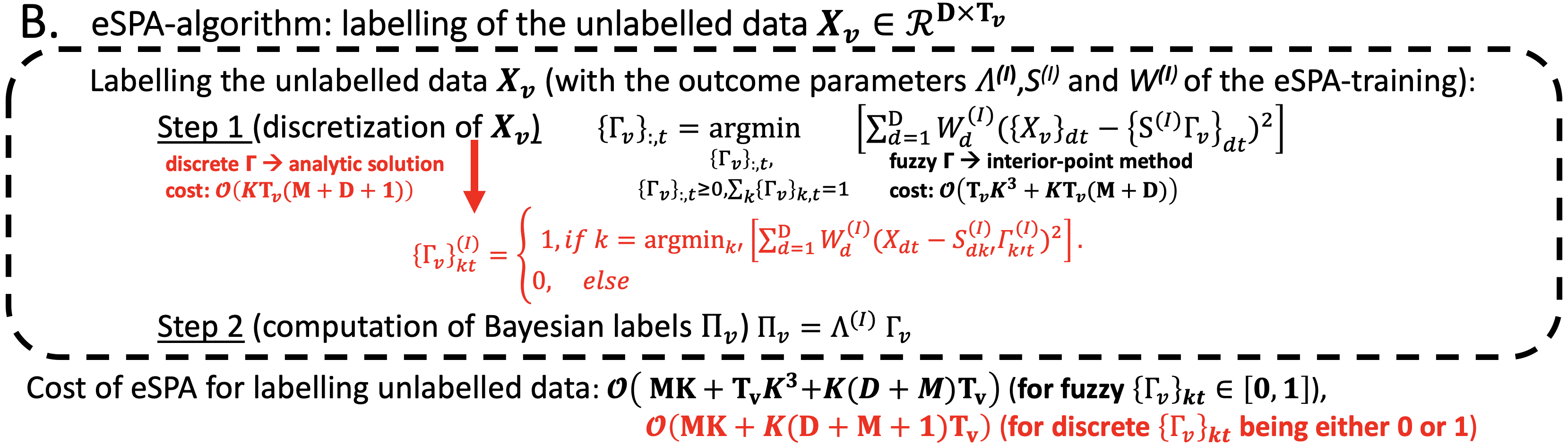} 
 \end{center}
   \caption{ {\bf Graphical explanation of the eSPA algorithm}: 
   (A) eSPA model training in the fuzzy case (black) and in the discrete case (red); (B) eSPA labelling of the unlabelled data (e.g., during the model validation step).
   }
    \label{fig:algorithm_and_sketch2}
\end{figure*}  

Defininging a scalar parameter $\epsilon_{CL}\geq0$, joint optimisation of the problems (\ref{eq:L_KL}) and (\ref{eq:L_e}) is formulated as a minimization of the joint functional $L$:
\begin{eqnarray}
\label{eq:L}
L&=&\frac{1}{T}\sum_{d=1}^DW_d\sum_{t=1}^T(X_{dt}-\left\{S\Gamma\right\}_{dt})^2 +\frac{\epsilon_e}{D}\sum_{d=1}^DW_d\log(W_d)-\frac{\epsilon_{CL}}{TM}\sum_{m=1}^M\sum_{t=1}^T\Pi_{mt}\log\left(\sum_{k=1}^K\Lambda_{mk}\Gamma_{kt}\right),\nonumber\\
&&\text{such that } \Gamma, \Lambda \text{ are stochastic matrices and } W \text{ is a probability vector,}
\end{eqnarray}
 where the scalar parameter  $\epsilon_{CL}\geq0$ regulates the relative importance of matching the relations between the labels and the boxes with a Bayesian network $\Lambda$ (see Fig.~2A). Setting $\epsilon_{CL}=0$ results in a solution of the unsupervised discretization and feature selection problem (\ref{eq:L_e}) only. Changing $\epsilon_{CL}$ controls the relative importance of the supervised labelled learning problem part compared to the unsupervised part of the problem. Graphic representation of the  problem (\ref{eq:L}) is given in the Fig.~2B.
 
 Given the data matrices $X$ and $\Pi$, for fixed values of $K, \epsilon_{CL}, \epsilon_e$ a solution of the optimisation problem (\ref{eq:L}) can be found numerically, by means of the entropy-based Scalable Probabilistic Approximation algorithm (eSPA) described in the Fig.~3. Characteristics of this algorithm and  the mathematical properties of the optimal solution for (\ref{eq:L}) are summarised in the following Theorem. Its proof is provided in the manuscript Appendix.  

\paragraph{Theorem:}\emph{ optimal solution of (\ref{eq:L}) is given by a segmentation $\Gamma$ that is piecewise-linear in the feature space $X$. For given $X\in\mathcal{R}^{D\times T}, \Pi\in\mathcal{R}^{M\times T},K, \epsilon_{CL}, \epsilon_e$ this optimal solution can be approximated by the monotonically-convergent eSPA algorithm (see Fig.~3). Iteration cost of the eSPA algorithm scales as $\mathcal{O}\left(KT(D+M)+KM+T(K+M+D)+D\log(D)\right)$ if the segmentation is discrete (i.e., if $\Gamma_{kt}$ is either 0 or 1, $\forall k,t$) and scales as $\mathcal{O}\left(KT(D+M)+K^3(M^3+T+1)+D\log(D)\right)$ if the segmentation is fuzzy (i.e., if $\Gamma_{kt}\in\left[0,1\right]$, $\forall k,t$). SPA algorithm and K-means clustering, having the same lead order computational complexity scaling,  are providing suboptimal box discretizations of $X$ compared to the discretizations obtained with the eSPA algorithm on the same data.  }   \\
 
\subsection*{{\emph{Numerical comparison of eSPA to the ML methods}}}
Next, eSPA algorithm is applied to the toy model example 1 from the Fig.~1A. The three user-defined tuneable hyper-parameters $K$, $\epsilon_{e}$ and $\epsilon_{CL}$ from (\ref{eq:L}) are selected from all possible combinations in ranges $K=\left[2,3,4,\dots,20\right]$,  $\epsilon_{e}=\left[0,10^{-5},10^{-4},\dots,10^{-1}\right]$ and $\epsilon_{CL}=\left[0,10^{-5},10^{-4},\dots,10^{-1}\right]$. The same multiple cross-validation procedure for training, validation and hyper-parameter selection is used for the eSPA and for the other ML models in Fig.~1. As can be seen from the middle panel of the Fig.~1A, eSPA achieves an almost perfect distinction between the two labelled sets on the validation data  (with AUC close to $1.0$) already with $K=3$ spatial boxes in feature space, correctly detecting the two original gene feature dimensions that are only relevant for a distinction between the classes. Contour plots of the three  piecewise-linear boxes in the two identified feature dimensions are shown in the right panel of the Fig.~1A, capturing the sandwich-like geometrical structure distinguishing the three Gaussian distributions in the feature space.    
\begin{figure*} 
\begin{center} 
     \includegraphics[width=1.0\textwidth]{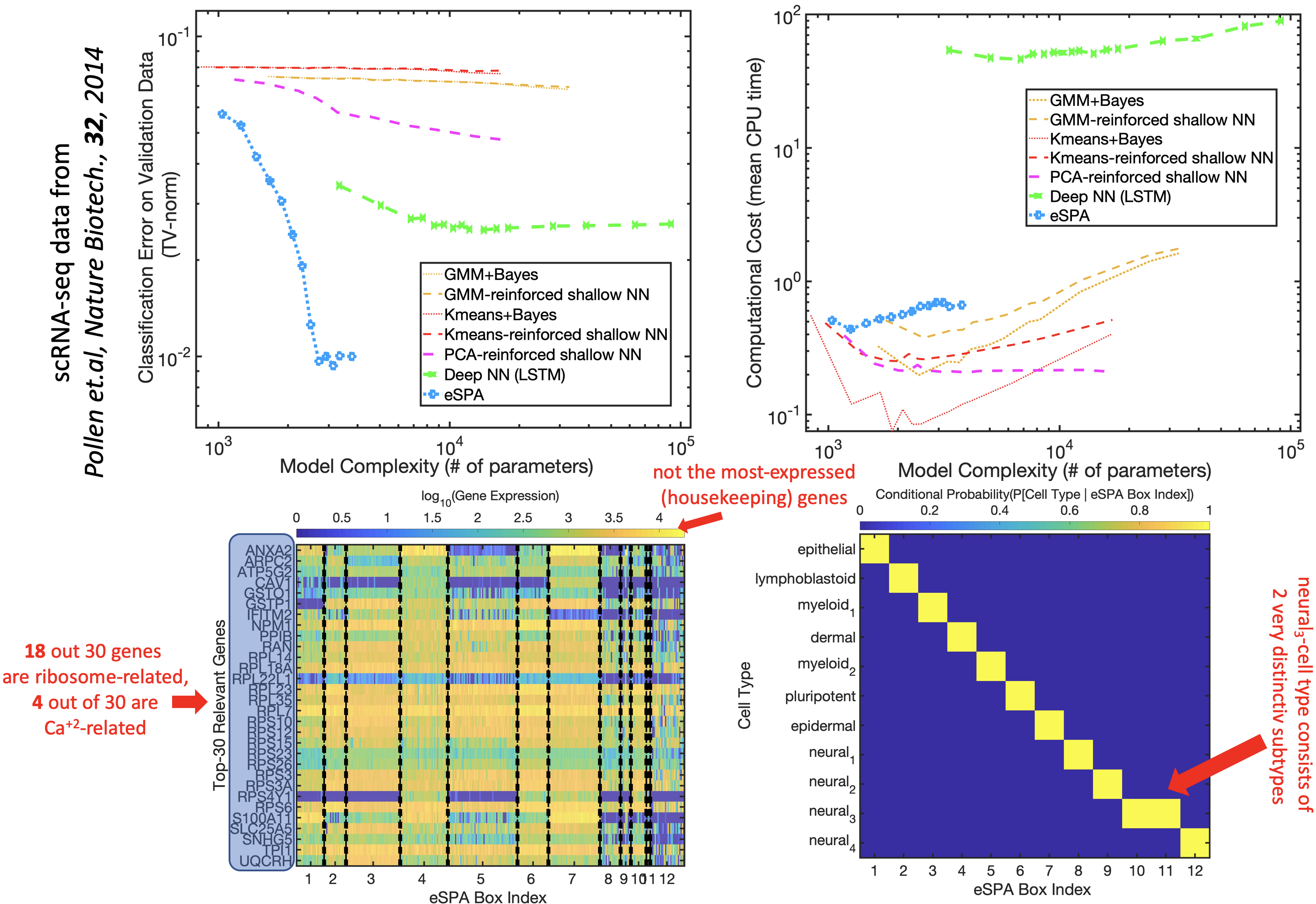} 
\end{center}
   \caption{ {\bf Comparison of ML and eSPA tools for the supervised classification of the single cell RNA gene expressions profiles data from \cite{pollen14}.}}  
    \label{fig:rnaSeq}
\end{figure*}  

 Fig.~1B shows a comparison of the eSPA to common ML methods on the non-Gaussian toy example 2 (see the left panel of the  Fig.~1B). Again, whereas the multiple cross-validation trials reveal that the  common tools fail to detect a correct classification rule (with all of the AUC values being close to 0.5 on validation data - and with no improvement when  increasing model complexity), eSPA detects the correct two feature dimensions and the correct classification rule with an AUC close to 1.0 when $K=5$, by decomposing the feature space in five piecewise-linear geometric boxes (see the right panel of the Fig.~1B).
 
 Results of comparing eSPA to LSTM and SVM on the generic classification problems of different feature dimensions $D$ and statistics sizes $T$ are shown in the Fig.~1C. They reveal no overfitting barrier for eSPA, as well as a robust identification of the correct classification rule also in the "small data challenge"  domain (where $T$ is small and $D$ is large). This result is obtained without a combinatorial search through all possible combinations of $D_{max}$-dimensional subsets in a $D$-dimensional feature space - and  at a fraction of the LSTM computational cost (see a right panel of the Fig.~1C).
 
 Fig.~4 shows results for the supervised classification of single cell RNA-sequencing data from a human brain sample  \cite{pollen14}. These data captures the gene expression levels  for over 20'000 various genes ($D=23'730$) in 301 brain tissue cells ($T=301$). Every cell is labelled according to its cell type determined manually and belongs to a one of the eleven cell type categories (for example, an "epithelial cell", a "neural$_1$ cell", etc.). Common single cell RNAseq-data processing pipelines from bioinformatics like the SEURAT (\url{https://github.com/satijalab/seurat/wiki/Assay}) perform an initial feature extraction (based, for example, on PCA and the non-linear dimension reduction methods like t-SNE ), followed by the clustering (K-means, density-based, etc.) of these reduced data into groups.   Results of applying eSPA to these data (see the upper left panel of the Fig.~4) reveal that in comparison to the common bioinformatics and ML methods, eSPA provides a thirty-fold improvement of cross-validated cell classification quality when compared to  these common bioinformatics methods - and a seven-fold improvement when compared to the multiply cross-validated deep learning classification based on the LSTM. This is achieved by a fraction of model complexity (see the x-axis of the upper left panel of the Fig.~4) and with the computational cost that is fractional to deep learning (see the upper right panel of the Fig.~4). Results of the eSPA can be interpreted biologically, indicating the genes that are most relevant for a single cell distinction (see the lower left panel Fig.~4). The Bayesian network matrix $\Lambda$ is shown in the lower right panel of Fig.~4, indicating that every cell type class in the validation data can be deterministically assigned to one of the geometric boxes obtained in the unsupervised part of the problem (\ref{eq:L}). It also shows that the cell type "neural$_3$" consists of two distinct geometric boxes - indicating that the cell type "neural$_3$" consists of two cell types that differ significantly with respect to their gene expression profiles. 
\section*{{\emph{Discussion}}} 
Using the simple generic test systems from bioinformatics, a joint comparison of classification quality and the computational cost for the popular classes of ML methods was performed.  It was shown that for these simple generic examples common methods like the deep learning LSTM and the SVM posses statistically highly-significant (p-value=$6\cdot10^{-7}$ for LSTM) overfitting barriers. One of the particular problems induced by the existence of these overfitting boundaries is the restrictive upper bound for the maximal number of robust feature dimensions $D_{max}$. As was shown for the toy model example 1, in the realistic case of $T=80, D=500$ (motivated by the biological data from \cite{lan19}), $D_{max}=5$ and the total number of $2.5\cdot10^{11}$ feature combinations has to be checked with the deep learning LSTM  classifier - resulting in around 1 Mio EUR in electricity costs only for a complete mining of the data and the robust identification of the correct classification rule for this single data set (see the lower panel of the Fig.~1C). 

 Presented results show that the entropy-optimising generalization of the recently-introduced SPA algorithm \cite{gerber20} allows to breach this overfitting barrier - without deploying any preliminary data-reduction steps like PCA or t-SNE and providing reliable results also when $T<D$ (see Fig.~1). eSPA identifies the correct rule without a combinatorial overhead and in a single analysis run - with an overall electricity cost of $10^{-6}$ EUR. This makes a total difference of twelve orders of magnitude in terms of cost when compared to the LSTM on the same problem.  These results indicate a possibility to treat the overfitting in deep learning applications by incorporating the entropic and geometric ideas like the ones used in the derivation of the eSPA algorithm presented in this paper.

\section*{{\emph{Materials and Methods}}} 
Code generating the data used in the Fig.~1 is  available under   
\url{https://www.dropbox.com/sh/sbhlp7r6zvbytg6/AAAIzPBKhgvZetZm3RLxPMQSa?dl=0}. Distribution variance parameter $sigma$ was chosen as $5.0$, $T=500$. Deep learning applications and the other common ML methods were trained and validated using the same commercial "Deep Learning" and the "Machine Learning" toolboxes by MathWorks. The code can be shared upon the email request.   
\section*{{\emph{Appendix}}} 
\subsubsection*{Proof of the Theorem}
\paragraph{a) Piecewise-linearity of the entropy-optimal box segmentation.} Let $S$ and $W$ be fixed with their (unknown) optimal values for (\ref{eq:L}). Then, solution of (\ref{eq:L}) for $\Gamma$ is equivalent to a minimization with respect to $\Gamma$ of the transformed problem 
\begin{eqnarray}
\label{eq:L_tilde}
\tilde{L}&=&\frac{1}{T}\sum_{d=1}^D\sum_{t=1}^T(\tilde{X}_{dt}-\left\{\tilde{S}\Gamma\right\}_{dt})^2 ,\\
&&\text{such that } \Gamma \text{ is a stochastic matrix,} \nonumber
\end{eqnarray}
 where $\tilde{X}=\text{diag}(W^{1/2})X, \tilde{S}=\text{diag}(W^{1/2})S$ and $\text{diag}(W^{1/2})$ is the diagonal matrix with a vector of square roots of the elements of $W$ on the diagonal. According to the Lemma 14 from \cite{gerber20} (available at \url{https://advances.sciencemag.org/content/advances/suppl/2020/01/27/6.5.eaaw0961.DC1/aaw0961_SM.pdf}), solution $\Gamma$ of (\ref{eq:L_tilde}) is a piecewise-linear function of the transformed features $\tilde{X}$. Hence, it is a piecewise-linear function of the original features $X$ (since  $\tilde{X}=\text{diag}(W^{1/2})X$ is a linear transformation of $X$). 
\paragraph{b) Monotonic convergence of the eSPA algorithm.} Every iteration of eSPA (see Fig.~3A) consists of four optimization steps, when each one of the four variables ($\Gamma, S, W, \Lambda$) is treated as an optimisation variable whereas the three other variables are fixed. It leads to a monotonic minimisation of the $L$ function values in every of the four steps - since the problems with respect to $W,  \Gamma$ and $\Lambda$ are convex optimisation problems on a simplex domain of linear constraints. The unconstrained convex optimisation problem with respect to $S$ is analytically solvable. Hence,  (\ref{eq:L}) is minimised monotonically in every eSPA iteration. This monotonic sequence will converge to a (local) minimum of (\ref{eq:L}) since $L\geq0$.
\paragraph{d) Iteration complexity of eSPA in the fuzzy case.} Monotonic minimisation of $L$ in the $W-$, $\Gamma-$ and $\Lambda-$ steps of the eSPA 
 algorithm (see Fig.~3A) can be achieved  with one iteration step of the interior-point method of convex optimization each \cite{nocedal06}. Computation of the $S$-step of eSPA can be performed analytically, resulting in a least-squares solution with a cost of $\mathcal{O}\left(K^3+KDT\right)$. Estimates of the iteration cost scalings for the single interior point iterations in every eSPA-step are given in the Fig.~3A. Sum of the cost scalings in the four steps results in the overall iteration complexity estimate $\mathcal{O}\left(KT(D+M)+K^3(M^3+T+1)+D\log(D)\right)$   
\paragraph{c) Iteration complexity of eSPA in the discrete case.}  If  $\Gamma_{kt}$ is either 0 or 1 ($\forall k,t$) then (\ref{eq:L}) is equivalent to a following minimization problem:
 \begin{eqnarray}
\label{eq:L_discr}
L_{discr}&=&\frac{1}{T}\sum_{d=1}^DW_d\sum_{k=1}^K\sum_{t=1}^T\Gamma_{kt}(X_{dt}-S_{dk})^2 +\frac{\epsilon_e}{D}\sum_{d=1}^DW_d\log(W_d)\nonumber\\
&&-\frac{\epsilon_{CL}}{TM}\sum_{m=1}^M\sum_{t=1}^T\Pi_{mt}\sum_{k=1}^K\Gamma_{kt}\log\left(\Lambda_{mk}\right),\nonumber\\
&&\text{such that } \Gamma, \Lambda \text{ are stochastic matrices and } W \text{ is a probability vector.}
\end{eqnarray}
Deploying the method of Lagrange multipliers the $\Gamma$- and the $\Lambda$-steps can be solved analytically. Analytical solutions and their computational cost estimates are marked red in the Fig.~3A. This reduces the overall eSPA iteration cost to $\mathcal{O}\left(KT(D+M)+KM+T(K+M+D)+D\log(D)\right)$. 
If the optimal $\Gamma$ is fuzzy, minimization  of (\ref{eq:L_discr}) will provide an approximate solution, since due to Jensen's inequality $L\leq L_{discr}$, meaning that the problem (\ref{eq:L_discr})- having a more favourable linear iteration complexity scaling - provides an upper-bound approximation of the problem (\ref{eq:L}).  
\paragraph{e) Sub-optimailty of SPA and K-means.} For $\epsilon_{CL}=0$ and $\epsilon_e\to\infty$, $W_i\to1/D$ and a solution of  (\ref{eq:L}) converges to the solution of the SPA problem (\ref{eq:L_SPA}). Hence,  eSPA provides an equivalent solution when compared to SPA if the eSPA problem  (\ref{eq:L})  is solved imposing an additional equality constraint $W_i=1/D$. Hence $L\leq L_{SPA}$ and SPA solutions are sub-optimal with respect to the solutions of eSPA.  Suboptimality of K-means with respect to SPA follows from the Corollary 1 from  \cite{gerber20} (available at \url{https://advances.sciencemag.org/content/advances/suppl/2020/01/27/6.5.eaaw0961.DC1/aaw0961_SM.pdf}). Hence, due to the transitivity K-means is suboptimal with respect to eSPA.  

 \paragraph{Acknowledgement}
 This work was supported by the "Emergent AI Center" of the JGU Mainz through a project "How emergent is the brain?" (financed by the Carl-Zeiss-Stiftung) and by the Mercator Fellowship in the DFG Collaborative Research Center 1114 "Scaling Cascades in Complex Systems". 
   \paragraph*{Competing Interests}
The author declares that he has no competing interests.

%
\bibliographystyle{naturemag}
\bibliography{bib_NatCom}

\end{document}